\documentclass{article}
\pdfoutput=1

\usepackage[preprint]{neurips_2024}

\usepackage[utf8]{inputenc} 
\usepackage[T1]{fontenc}    
\usepackage{hyperref}       
\usepackage{url}            
\usepackage{booktabs}       
\usepackage{amsfonts}       
\usepackage{nicefrac}       
\usepackage{microtype}      
\usepackage{xcolor}         
\usepackage{multirow}
\usepackage{graphicx}
\usepackage{amsmath}
\usepackage{makecell}
\usepackage{wrapfig}
\usepackage{array}
\usepackage{subcaption}

\title{Multi-level Interaction Modeling \\for Protein Mutational Effect Prediction}

\author{
Yuanle Mo\textsuperscript{1}\thanks{Equal contribution}\;\,\thanks{Work was done while Yuanle Mo was a research intern at AIR.}\;\,, 
Xin Hong \textsuperscript{2}\footnotemark[1]\;\,, 
Bowen Gao\textsuperscript{2}, 
Yinjun Jia\textsuperscript{3}, 
Yanyan Lan\textsuperscript{2}\thanks{Correspondence to \texttt{lanyanyan@air.tsinghua.edu.cn}}  \\
\\
\textsuperscript{1}School of Information and Software Engineering, UESTC \\
\textsuperscript{2}Institute for AI Industry Research (AIR), Tsinghua University \\
\textsuperscript{3}School of Life Sciences, Tsinghua University \\
}

\begin{document}

\maketitle

\begin{abstract}

Protein-protein interactions are central mediators in many biological processes. Accurately predicting the effects of mutations on interactions is crucial for guiding the modulation of these interactions, thereby playing a significant role in therapeutic development and drug discovery. Mutations generally affect interactions hierarchically across three levels: mutated residues exhibit different \textit{sidechain} conformations, which lead to changes in the \textit{backbone} conformation, eventually affecting the binding affinity between \textit{proteins}. However, existing methods typically focus only on sidechain-level interaction modeling, resulting in suboptimal predictions. In this work, we propose a self-supervised multi-level pre-training framework, ProMIM, to fully capture all three levels of interactions with well-designed pre-training objectives. Experiments show ProMIM outperforms all the baselines on the standard benchmark, especially on mutations where significant changes in backbone conformations may occur. In addition, leading results from zero-shot evaluations for SARS-CoV-2 mutational effect prediction and antibody optimization underscore the potential of ProMIM as a powerful next-generation tool for developing novel therapeutic approaches and new drugs.

\end{abstract}

\section{Introduction}
\label{sec:intro}

Protein–protein interactions (PPIs) are central mediators in many biological processes\cite{rual2005towards,acuner2011transient,bludau2020proteomic}, including signal transduction, metabolic regulation, structural maintenance, etc. Given their critical roles, modulating PPIs has emerged as a strategic approach in both therapeutic development and drug discovery. A common approach to modulating PPIs involves introducing specific mutations that either strengthen or weaken these interactions, such as enhancing the effect of a neutralizing antibody against a virus and disrupting harmful PPIs that drive pathological conditions in cancer \cite{liu2022neutralizing,carels2023strategy}. Traditional methods understand the mutational effects through experimental assays or computational approaches. However, experimental assays have limitations in their scalability, and the choice of the measured phenotype highly influences the relevance of results to organism fitness and physiology \cite{hopf2017mutation}. These limitations lead to the development of computational methods, whose target is to predict the change in binding free energy ($\Delta\Delta G$), which can be categorized into energy-based and statistics-based methods. The former methods typically use physical energies and statistical potential to predict the $\Delta\Delta G$ but often suffer from computational burden and poor generalization ability \cite{schymkowitz2005foldx,dehouck2013beatmusic,xiong2017bindprofx}. The latter ones have better scalability and rapid predictive capability but rely on handcrafted features as inputs, which makes it hard to fully capture intricate interactions between proteins \cite{pires2016mcsm,li2016mutabind}.

\begin{wrapfigure}{r}{0.5\textwidth}

\vspace{-15pt}

\centering
\includegraphics[width=\linewidth]{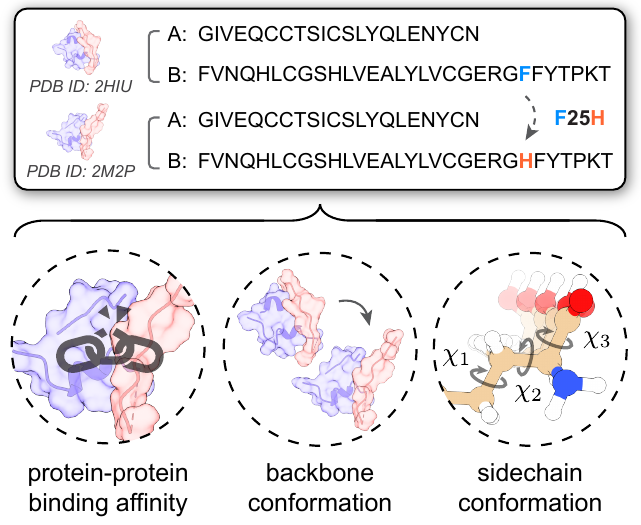}
\caption{Three levels of mutational effects on protein-protein interaction.
}
\label{fig:mutation_effects}
    
\vspace{-10pt}

\end{wrapfigure}

Recent advances in AI4Science area, especially for protein structure prediction~\cite{senior2020improved,jumper2021highly,abramson2024accurate}, show promise for solving mutational effect prediction with deep learning. Compared to the massive existing protein structure data, protein mutation data with labeled binding affinity changes is extremely scarce. Therefore, the challenge of mutation prediction lies in leveraging mutation-unlabeled data, e.g. protein complex structures, through pre-training to learn mutation-related features. Previous methods noticed that mutation mainly changes sidechain conformations~\cite{liu2021deep,luo2023rotamer,liu2023predicting}, so they pre-train models via optimizing sidechain conformations and achieve great performance by further tuning on mutation data. However, we argue that it is incomplete to only model sidechain conformations, since the change of sidechain conformations cannot fully characterize the variations of PPIs upon mutation. For instance, the conformational flexibility of glycine sidechains and the rigidity of proline sidechains play an important role in defining the backbone flexibility. Mutations from and to these two amino acids not only cause changes in the sidechain conformations, but can also lead to large structural effects on the protein backbone, and influence the binding strength between proteins \cite{rodrigues2019mcsm}. Generally, as the Figure~\ref{fig:mutation_effects} shown, the mutational effect on protein-protein binding can be simplified and modeled as a hierarchical process occurring at three related levels: mutated residues exhibit different \textit{sidechain} conformations, which lead to changes in the \textit{backbone} conformation, e.g. flexibility, eventually affecting the binding affinity between \textit{proteins} \cite{teng2010structural,xiong2022implications,koseki2023topological}.



To fully model all three levels of mutational effects on protein-protein interactions, we propose a self-supervised multi-level pre-training framework, \textbf{M}ulti-level \textbf{I}nteraction \textbf{M}odeling for \textbf{Pro}tein mutational effect prediction (ProMIM). In the protein-level interaction modeling (PIM), we formulate the protein binding task as a matching problem, i.e. whether two given proteins bind to each other or not. Inspired by recent studies in the image-text \cite{li2021align,radford2021learning} and protein-ligand matching \cite{gao2024drugclip,gao2023self}, we optimize protein-protein matching with contrastive learning to predict whether in-batched proteins are the binding partners. In the finer backbone-level interaction modeling (BIM), the most crucial aspect is the ability to be aware of the changes in the backbone conformation caused by mutation. ProMIM improves the model's sensitivity on backbone conformation change via predicting the distance map between protein binders, which is a commonly used structure prediction objective \cite{guo2022prediction,zhu2023unified}. Finally, at the forefront of protein-protein interaction, particularly in sidechain-level interaction modeling (SIM), we build upon previous work by RDE~\cite{luo2023rotamer}, utilizing flow-based methods to model the distributions of rotamers (i.e., sidechain conformations). This allows us to estimate entropy loss on the binding interface, which contributes to determining the free energy change caused by mutation.

Due to ProMIM's comprehensive modeling of interactions at different levels, it surpasses previous methods on the SKEMPI2 dataset \cite{jankauskaite2019skempi} that only model sidechain conformations (Section~\ref{sec:skempi2}), especially on mutations most likely to affect backbone conformations (Section~\ref{sec:sidechain_verify}). Ablation studies (Section~\ref{sec:ablation}) demonstrate the indispensability of all three levels of interaction modeling. Additionally, it shows a significant advantage of the self-supervised design philosophy in ProMIM, with an improved performance of 8.29\% on the per-structure Spearman correlation coefficient compared to the model learning from scratch on SKEMPI2. This is because ProMIM can learn from a larger scale of protein data, PPIRef50K and PDB-REDO, which are more than 370 times larger than the mutation dataset SKEMPI2. Furthermore, results from zero-shot evaluations for SARS-CoV-2 mutational effect prediction \cite{starr2022shifting} (Section~\ref{sec:SARS ddg}) and antibody optimization \cite{shan2022deep} (Section~\ref{sec:antibody}) demonstrate the great generalization ability of ProMIM, indicating that ProMIM has the potential to become a next-generation tool for developing new therapies and drugs.

In summary, our contributions are as follows. Firstly, we propose a novel self-supervised multi-level pre-training framework, ProMIM, which fully captures three levels of interactions with well-designed pre-training objectives. Secondly, extensive experiments demonstrate that our method achieves state-of-the-art performance on the SKEMPI2 dataset, particularly surpassing previous methods on mutations more likely to change backbone conformations and overall binding affinity. Thirdly, the significant generalization ability shown in zero-shot evaluations of SARS-CoV-2 underscores the potential of ProMIM as a powerful tool for developing novel therapeutic approaches and new drugs.





\section{Related Work}

\subsection{Protein Mutational Effect Prediction}

Traditional methods for $\Delta\Delta G$ prediction broadly fall into two categories: energy-based and statistics-based methods. Energy-based methods utilize physical energies, such as van der Waals, electrostatic potential and hydrogen bonding for estimation \cite{schymkowitz2005foldx,alford2017rosetta,barlow2018flex}, while statistics-based methods use handcrafted features such as motif properties and evolutionary conservation to predict the mutational effects on binding \cite{pires2016mcsm,li2016mutabind}. Both of these approaches highly rely on human expertise, and fail to capture intricate interactions between proteins.

In contrast, deep learning-based methods have shown better promise, which can be categorized as supervised and pre-training-based models. The supervised methods extract features from wild-type and mutant complexes, and directly predict $\Delta\Delta G$ \cite{shan2022deep} using labeled data in an end-to-end manner. Nevertheless, their development is hindered by the scarcity of labeled protein mutation data. Pre-training-based methods attempt to mitigate this challenge with various pre-training objectives. Some employ masked modeling on protein structures, training models to classify masked amino acids. The predicted probabilities of amino acid types before and after mutation are then used to estimate $\Delta\Delta G$, but the correlation between residue type probabilities and changes in binding free energy tends to be mild \cite{hsu2022learning,bushuiev2023learning}. Other methods leverage protein representations pre-trained with sidechain modeling objectives to capture mutation-related features, achieving state-of-the-art performance for $\Delta\Delta G$ prediction \cite{liu2021deep,luo2023rotamer,liu2023predicting}. However, these methods only incorporate sidechain-level interaction modeling, so their effectiveness can be impaired when mutations lead to significant changes in protein backbone conformations and overall binding affinity. To this end, our approach fully models PPI across different levels to facilitate a comprehensive understanding of mutational effects on binding.

\subsection{Protein-Protein Interaction Modeling}

Protein-protein interaction modeling has been studied at each level for decades. Since mutations hierarchically impact interactions across different levels, comprehensive modeling of each level is required to fully characterize the mutational effects on PPIs.

Protein-level interaction modeling typically refers to the task of identifying potential protein binding pairs. Experimental methods are often costly and time-consuming \cite{fields1994two,burckstummer2006efficient}, so recent computational approaches use deep learning to predict protein interactions. These approaches extract informative features from different modalities of protein data to predict interactions \cite{zhang2019sequence,zhao2023semignn,bryant2022improved,gao2023hierarchical,wu2024mape}. Some latest works also explore correlations within PPI networks to infer unknown interactions \cite{yang2020graph,lv2021learning}. In the context of mutational analysis, predicting protein interactions is challenging, since a small number of mutations on proteins can lead to significantly different binders \cite{hashimoto2010mechanisms,li2014predicting}. Therefore, our method adopts a contrastive manner, which has been successfully verified in multimodal learning \cite{li2021align,radford2021learning}, to model protein-level interactions in order to emphasize learning the relationship between inter-protein differences and resulting binding patterns.


The representative task for backbone-level interaction modeling is protein docking, which involves predicting the 3D structures of complexes from unbound states. Classical docking software typically suffers from intensive computational burden. \cite{schindler2017protein,sunny2021fpdock,torchala2013swarmdock,vakser2014protein}. Deep learning-based methods have emerged as better alternatives, capturing evolutionary constraints and geometric features from growing protein data \cite{ganea2021independent,evans2021protein,wang2024injecting}. Recent approaches focus on rigid-body protein-protein docking, predicting rotation and translation to attain the bound structure \cite{ketata2023diffdock,yu2024rigid,wu2023neural}. Nevertheless, predicting rigid transformations cannot fully capture the complicated spatial changes in inter-molecular distances caused by mutations. Consequently, for backbone-level interaction modeling, our method differs from previous docking approaches by predicting the relative distance between proteins, which implicitly models the flexibility of backbone conformations.

Sidechain-level interaction modeling focuses on predicting rotamers given the protein backbone structures. Traditional methods for rotamer prediction operate by minimizing the energy function across a pre-defined rotamer library \cite{xu2006fast,krivov2009improved,huang2020faspr,leman2020macromolecular}. Recent methods employ various deep learning techniques such as 3D convolution network \cite{misiura2022dlpacker}, SE(3)-Transformer \cite{mcpartlon2022end} and diffusion models \cite{zhang2024diffpack,liu2023predicting}, to achieve more accurate rotamer prediction. Among them, several approaches utilize sidechain modeling to predict mutational effects and exhibit superior performance, including RDE \cite{luo2023rotamer} employing a flow-based generative model to estimate the probability distribution of rotamers, as well as SidechainDiff \cite{liu2023predicting} using a Riemannian diffusion model to learn the generative process of rotamers. These methods inject mutation-related knowledge into pre-trained protein representations via sidechain-level interaction modeling.

\section{Method}
In this section, we first present the notations used in the paper and provide a schematic description of our model architecture (Section \ref{sec:notation&model}). Second, we introduce three pre-training objectives that model interactions at different levels (Section \ref{sec:pre-training obj}). Third, we demonstrate how to predict $\Delta\Delta G$ values with the pre-trained models (Section \ref{sec:mutational effect prediction}). For training details, please refer to the Appendix \ref{apd:imp details}.

\subsection{Notations and Model Overview}
\label{sec:notation&model}

\textbf{Notations} In protein-protein interaction, we denote a complex as $\mathbf{c}\in\mathbb{A}^n$, where $n$ represents the number of residues, and $\mathbb{A}$ is the alphabet of amino acids $\{1, \dots, 20\}$. Complexes with two binding proteins consist of a receptor protein and a ligand protein, represented as disjoint sets of indices $\mathcal{P}^r$ and $\mathcal{P}^l$, respectively.
Each residue in the complex is characterized using its type $a_i\in \mathbb{A}$, position $\boldsymbol{p}_i\in\mathbb{R}^3$, orientation $\boldsymbol{O}_i\in SO(3)$, and the sidechain torsion angles $\boldsymbol{\chi}_i=(\chi_i^{(k)})_{k=1}^t$ where $\chi_i^{(k)}\in[0,2\pi)$ and $t$ denotes the number of torsion angles $(i\in\{1\dots n\})$.

\begin{figure}[t]
\centering
\includegraphics[width=\linewidth]{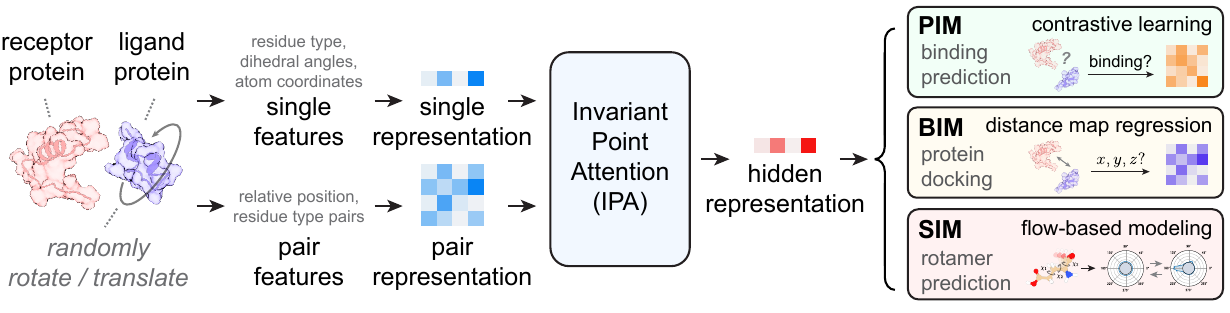}
\caption{The overview framework of the ProMIM.}
\label{fig:ppi_network}
\end{figure}

\textbf{Model Overview}
The overall framework of ProMIM is illustrated in Figure \ref{fig:ppi_network}. In the early stage of attaining residue-wise representations, we follow RDE \cite{luo2023rotamer} to capture information about each residue and its structural context. Specifically, given a complex, two protein binders are randomly assigned as the receptor protein and ligand protein. To simulate an unbound state, the receptor protein is kept at its initial location, while random transformations, including rotation and translation, are performed to the ligand protein to change its location. Next, single and pair features are extracted from the complex. Single features include the residue type, backbone dihedral angles, and local atom coordinates, while the pair features include the relative position and residue type pairs. We denote the single embeddings and pair embeddings as $\boldsymbol{e}_{i}$ and $\boldsymbol{z}_{ij}$ respectively, where $i,j\in\{1\dots n\}$. These two features are then transformed into hidden representations $\boldsymbol{h}_i$ with the Invariant Point Attention Module (IPA), introduced by AlphaFold2 \cite{jumper2021highly}. The resulting residue-wise representations $\boldsymbol{h}_i$ are then employed to conduct the subsequent interaction-aware pre-training process across three levels.

\subsection{Three Levels of Interaction Modeling}
\label{sec:pre-training obj}
In this section, we introduce how ProMIM designs pre-training objectives for modeling each level of interaction, to fully capture the protein mutational effects.

\textbf{Protein-level Interaction Modeling}
Mutation can change the binding relation between protein pairs, i.e. binding affinity. In this paper, we formulate PIM as a matching problem to capture this binding relation in a coarse grain, namely whether a pair of proteins should bind together or not, which reduces the need for data containing affinity labels. Matching proteins is challenging since proteins with highly similar sequences can participate in different interactions, and even a small number of amino acid mutations can lead to different interaction partners \cite{hashimoto2010mechanisms,li2014predicting}. This is similar to the problem of matching documents and images \cite{li2021align,radford2021learning} in multimodal learning, where the superficial differences between words and pixels are not the matching key, but the underlying semantics are. Advanced works in this field, such as CLIP \cite{radford2021learning}, inspire us to adopt the contrastive learning objective to model the protein-level interactions.
 

Specifically, for two proteins $\mathcal{P}^l$ and $\mathcal{P}^r$ belonging to a complex $\mathbf{c}$, we apply max-pooling to the residue-wise hidden representations $\boldsymbol{h}_i(i\in\{1\dots n\})$ obtained from the encoder to get a global structure representation for each protein, denoted as $\boldsymbol{H}^l$ and $\boldsymbol{H}^r$. Given a batch of protein complexes $\mathbf{c}_k(k\in\{1\dots N\})$ with batch size $N$, we attain global representations for each pair of proteins in the batch using our encoder. Subsequently, we combine them to form $N^2$ representation pairs $(\boldsymbol{H}^l_i,\boldsymbol{H}^r_j)$ where $i,j\in \{1\dots N\}$. When $i=j$ it is a positive pair, and when $i\neq j$ it is a negative pair. Two contrastive losses are introduced to facilitate the training process:
\begin{equation}
\mathcal{L}_k^l(\boldsymbol{H}_k^l,\{\boldsymbol{H}_i^r\}_{i=1}^N)=-\frac1N\log\frac{\exp(s(\boldsymbol{H}_k^l,\boldsymbol{H}_k^r)/\tau)}{\sum_i\exp(s(\boldsymbol{H}_k^l,\boldsymbol{H}_i^r)/\tau)},
\end{equation}
\begin{equation}
\mathcal{L}_k^r(\boldsymbol{H}_k^r,\{\boldsymbol{H}_k^l\}_{i=1}^N)=-\frac{1}{N}\log\frac{\exp(s(\boldsymbol{H}_k^l,\boldsymbol{H}_k^r)/\tau)}{\sum_i\exp(s(\boldsymbol{H}_i^l,\boldsymbol{H}_k^r)/\tau)},
\end{equation}
where $\tau$ is the temperature parameter that controls softmax distribution, and $s(\cdot)$ is a function to calculate the similarity score of a given representation pair. Here we adopt cosine similarity. Both two contrastive losses are utilized to identify the binding protein for a given one and enforce symmetry.

The final contrastive loss for a batch is formulated below:
\begin{equation}
\mathcal{L}_{PIM}=\frac12\sum_{k=1}^N(\mathcal{L}_k^l+\mathcal{L}_k^r),
\end{equation}

\textbf{Backbone-level Interaction Modeling}
At a finer level, mutation can change the backbone conformation of a complex, and predicting this conformation pertains to the protein docking problem. General docking methods typically predict a rotation and a translation on the input unbound structures to obtain their bound state \cite{ganea2021independent,wang2024injecting}. However, mutation can lead to complicated variations in inter-molecular distances, which cannot be adequately represented solely by rotation and translation. Therefore, we use the relative distance between protein pairs as the supervision signal for BIM, which can implicitly model backbone conformation such as its flexibility.



To be specific, given the residue-wise representations $\boldsymbol{h}_i(i\in\{1\dots n\})$ of a protein complex $\mathbf{c}$ formed by $\mathcal{P}^l$ and $\mathcal{P}^r$, we firstly transform them into a pairwise representation $\boldsymbol{h}_{lr}^{(0)}$ using a MLP, which serves as the attention bias added to the attention weight. Subsequently, the M-layer Transformer takes both residue-wise and pairwise representations as inputs and generates the final pairwise representation $\boldsymbol{h}_{lr}^{(M)}$. We denote the ground-truth distance map of the original complex structure in the bound state as $\mathcal{D}$, where each element $d_{ij}\in \mathcal{D}$ represents the distance between the $i$-th $C_{\alpha}$ atom in the ligand protein $\mathcal{P}^l$ and the $j$-th $C_{\alpha}$ atom in the receptor protein. The predicted distance map $\hat{\mathcal{D}}$ is obtained using a one-layer MLP to transform the final pairwise representation $\boldsymbol{h}_{lr}^{(M)}$. We regard the distance map prediction as a regression task and employ the Mean Square Error (MSE) loss in the training:
\begin{equation}
\mathcal{L}_{BIM}= \mathcal{L}_{MSE}(\mathcal{D}, \hat{\mathcal{D}}),
\end{equation}

\textbf{Sidechain-level Interaction Modeling}
When mutations occur, changes in protein-protein affinity and backbone structure actually stem from alterations in the nearby interaction environment~\cite{xu2023opus,eyal2003protein}, caused by the changes in sidechain conformations. Previous methods have well studied SIM by modeling rotamer distribution as the pre-training task and obtained great performance on mutational effect prediction. We adopt the flow-based modeling proposed by RDE \cite{luo2023rotamer}, where a conditional normalizing flow serves as a flexible method to model the complex probability density of rotamers by transforming it into a simple base density in an invertible manner \cite{durkan2019neural}.


Specifically, a rational quadratic spline flow \cite{durkan2019neural,rezende2020normalizing} is firstly utilized to model the distribution of a rotamer with one torsion angle, which is a piece-wise bijective function that contains $K$ pieces delimited by $K+1$ knots, with each piece formulated as follows:
\begin{equation}
\begin{aligned}f_k(x|x_{k,k+1},y_{k,k+1},\delta_{k,k+1})&=y_k+\frac{(y_{k+1}-y_k)\left[s_k\xi_k^2(x)+\delta_k\left(1-\xi_k(x)\right)\xi_k(x)\right]}{s_k+[\delta_{k+1}+\delta_k-2s_k]\left(1-\xi_k(x)\right)\xi_k(x)}, \\\mathrm{where~}s_k&=\frac{y_{k+1}-y_k}{x_{k+1}-x_k},\mathrm{and~}\xi(x)=\frac{x-x_k}{x_{k+1}-x_k}\quad(x\in[x_k,x_{k+1}]),\end{aligned}
\end{equation}

The bijective is parameterized by the coordinates and derivatives of the $K+1$ knots, denoted as $x_k$, $y_k$, and $\delta_k (k\in{1\dots K+1})$, and forms a strictly monotonically increasing function on $[0,2\pi]$, denoted as $f : [0,2\pi] \to [0,2\pi]$. These parameters are conditioned on the hidden representation $\boldsymbol{h}_i$ using neural network transformation, so the bijective can also be denoted as $f(x|\boldsymbol{h}_i)$. The uniform distribution $p_{z}(z)=\frac{1}{2\pi} (z\in[0,2\pi])$ is employed as the base distribution, and $f$ maps from the target rotamer distribution to the base distribution. According to the change-of-variable formula in the probability density function, we attain the target rotamer density as follows:
\begin{equation}
\log p(x|\boldsymbol{h}_i)=\log p_z\left(f(x)\right)+\log|f^{\prime}(x|\boldsymbol{h}_i)|=-\log2\pi+\log|f^{\prime}(x|\boldsymbol{h}_i)|
\end{equation}

Subsequently, coupling layers stack multiple bijectives to enable modeling distributions of rotamers with more torsion angles. The training process aims to minimize the negative log-likelihood of ground truth rotamers, where $n$ denotes the number of residues in the complex:
\begin{equation}
\mathcal{L}_{SIM}=-\frac{1}{n}\sum_{i=1}^n\log p(\boldsymbol{\chi}_i|\boldsymbol{h}_i),
\end{equation}

\subsection{Mutational Effect Prediction}
\label{sec:mutational effect prediction} 
ProMIM is pretrained on the PPIRef50K \cite{bushuiev2023learning} and PDB-REDO \cite{joosten2014pdb_redo} datasets. Please refer to the Appendix \ref{apd:imp details} for training details. Given a wild-type complex $\mathbf{c}_{wt}$ and its mutant complex $\mathbf{c}_{mt}$, we first obtain their residue-wise hidden representations $\boldsymbol{h}_i^{wt}$ and $\boldsymbol{h}_i^{mt}$ from the pre-trained ProMIM, which provides rich interaction knowledge across three levels. Next, we adopt the same encoder architecture in the pre-training stage for $\Delta\Delta G$ prediction. $\boldsymbol{h}_i^{wt}$ and $\boldsymbol{h}_i^{mt}$ are fused with the single embeddings using a one-layer MLP, following the same practice in RDE. A max-pooling layer is applied to the output of IPA module to attain global structure representations for the wild-type and mutant complexes, denoted as $\boldsymbol{H}^{wt}$ and $\boldsymbol{H}^{mt}$ respectively. Finally, these global representations are fed into another MLP to predict $\Delta\Delta G$. The MSE loss is used for training. 


\section{Experiments}


\subsection{Prediction of Mutational Effects on Protein-Protein Binding}
\label{sec:skempi2}

\paragraph{Experimental Configuration}
In order to ensure a fair comparison with baseline results, we adhere to the experimental protocols outlined in DiffAffinity\cite{liu2023predicting}. Specifically, a three-fold cross-validation approach is employed on the SKEMPI2 dataset, that is, the dataset is divided into three folds based on structural criteria, with each fold containing distinct protein complexes not found in the others. Two folds are utilized for training and validation purposes, while the third fold serves as the testing set.

\paragraph{Baselines}
We select a broad range of models as the baselines, including energy-based methods such as FoldX \cite{schymkowitz2005foldx}, Rosetta \cite{alford2017rosetta} and flex ddG \cite{barlow2018flex}; sequence-based methods such as ESM-1v \cite{meier2021language} and ESM2 \cite{lin2023evolutionary}; an unsupervised method ESM-IF \cite{hsu2022learning}; an end-to-end method DDGPred \cite{shan2022deep}; as well as pre-training methods consisting of ESM2* \cite{lin2023evolutionary}, RDE-Network \cite{luo2023rotamer} and DiffAffinity \cite{liu2023predicting}. Note that all pre-training methods have the same architecture for $\Delta\Delta G$ prediction but differ in pre-training objectives, which ensures a direct comparison among the effectiveness of pre-training objectives.

\paragraph{Metrics}
Pearson and Spearman correlation coefficients, root mean squared error (RMSE), and mean absolute error (MAE), as well as the area under the receiver operating characteristic (AUROC) are utilized to evaluate the performance. To calculate AUROC, we classify mutations into positive and negative effects based on the sign of $\Delta\Delta G$. Following the methodology outlined in DiffAffinity~\cite{liu2023predicting}, we also calculate two additional metrics: average per-structure Pearson correlation coefficient and average per-structure Spearman correlation coefficient, which are often of greater interest in practical applications. Specifically, we group mutations by complex structure, exclude groups with fewer than ten mutation data points, and calculate correlation coefficients for each complex separately.


\paragraph{Results}
Most baseline results come from the previous work~\cite{liu2023predicting}, while RDE-Network and DiffAffinity are implemented on our own. At first glance to Table \ref{tab:skempi2}, ProMIM achieves the best or second-best results across all metrics, indicating the effectiveness of ProMIM's training objectives. From the table, we can see that, in addition to the overall results, statistics are also provided separately for single-point and multi-point mutations. ProMIM's overall superior performance over RDE-Network and DiffAffinity is primarily due to its significant advantage in multi-point mutations, while its performance is comparable in single-point mutations. We believe this is because muti-point mutations have larger opportunity to affect the backbone conformation and overall binding affinity than single-point mutations, while these two types of interaction modeling are absent in RDE-Network and DiffAffinity. Table~\ref{tab:skempi2} also presents per-structure results, where correlation coefficients are computed for each structure individually and then averaged. This metric is more indicative of performance in practical applications and therefore receives attention. ProMIM's leading performance on this metric demonstrates its practical value and advantage.

\begingroup
\renewcommand{\arraystretch}{0.996} 

\begin{table}[t]
\small
\centering
\caption{Evaluation of $\Delta\Delta G$ prediction on the SKEMPI2 dataset. Results ranking first and second are highlighted in \textbf{bold} and \underline{underlined} respectively.}
\label{tab:skempi2}
\resizebox{\textwidth}{!}{
\begin{tabular}{cl|rrrrr|rr}
\toprule
& & \multicolumn{5}{c|}{Overall} & \multicolumn{2}{c}{Per-Structure} \\
Method & Mutations & Pearson  & Spearman & RMSE & MAE & AUROC & Pearson & Spearman \\
\midrule
\multirow{3}{*}{FoldX}
& all & 0.319 & 0.416 & 1.959 & 1.357 & 0.671 & 0.376 & 0.375 \\
& single & 0.315 & 0.361 & 1.651 & 1.146 & 0.657 & 0.382 & 0.360 \\
& multiple & 0.256 & 0.418 & 2.608 & 1.926 & 0.704 & 0.333 & 0.340 \\
\midrule
\multirow{3}{*}{Rosetta}
& all & 0.311 & 0.346 & 1.617 & 1.131 & 0.656 & 0.328 & 0.298 \\
& single & 0.325 & 0.367 & 1.183 & 0.987 & 0.674 & 0.351 & 0.418 \\
& multiple & 0.199 & 0.230 & 2.658 & 2.024 & 0.621 & 0.191 & 0.083 \\
\midrule
\multirow{3}{*}{flex ddG}
& all & 0.402 & 0.427 & 1.587 & 1.102 & 0.675 & 0.414 & 0.386 \\
& single & 0.425 & 0.431 & 1.457 & 0.997 & 0.677 & 0.433 & \underline{0.435} \\
& multiple & 0.398 & 0.419 & 1.765 & 1.326 & 0.669 & 0.401 & 0.363 \\
\midrule
\multirow{3}{*}{ESM-1v}
& all & 0.192 & 0.157 & 1.961 & 1.368 & 0.541 & 0.007 & -0.012 \\
& single & 0.191 & 0.157 & 1.723 & 1.192 & 0.549 & 0.042 & 0.027 \\
& multiple & 0.192 & 0.175 & 2.759 & 2.119 & 0.542 & -0.060 & -0.128 \\
\midrule
\multirow{3}{*}{ESM-IF}
& all & 0.319 & 0.281 & 1.886 & 1.286 & 0.590 & 0.224 & 0.202 \\
& single & 0.296 & 0.287 & 1.673 & 1.137 & 0.605 & 0.391 & 0.364 \\
& multiple & 0.326 & 0.335 & 2.645 & 1.956 & 0.637 & 0.202 & 0.149 \\
\midrule
\multirow{3}{*}{ESM2}
& all & 0.133 & 0.138 & 2.048 & 1.460 & 0.547 & 0.044 & 0.039 \\
& single & 0.100 & 0.120 & 1.730 & 1.210 & 0.541 & 0.019 & 0.036 \\
& multiple & 0.170 & 0.163 & 2.658 & 2.021 & 0.566 & 0.010 & 0.010 \\
\midrule
\multirow{3}{*}{ESM2*}
& all & 0.623 & 0.498 & 1.615 & 1.179 & 0.721 & 0.362 & 0.316 \\
& single & 0.625 & 0.468 & 1.357 & 0.986 & 0.707 & 0.391 & 0.342 \\
& multiple & 0.603 & 0.529 & 2.150 & 1.670 & 0.758 & 0.333 & 0.304 \\
\midrule
\multirow{3}{*}{DDGPred}
& all & 0.630 & 0.400 & \textbf{1.313} & \textbf{0.995} & 0.696 & 0.356 & 0.321 \\
& single & 0.652 & 0.359 & 1.309 & 0.936 & 0.656 & 0.351 & 0.318 \\
& multiple & 0.591 & 0.503 & 2.181 & 1.670 & 0.759 & 0.373 & 0.385 \\
\midrule
\multirow{3}{*}{RDE-Net}
& all & 0.654 & \underline{0.555} & 1.546 & 1.104 & \underline{0.749} & \underline{0.455} & \underline{0.424} \\
& single & 0.647 & 0.515 & 1.307 & 0.945 & 0.730 & \underline{0.456} & 0.428 \\
& multiple & \underline{0.647} & \underline{0.596} & \underline{2.005} & \underline{1.506} & \underline{0.806} & \underline{0.458} & \textbf{0.450} \\
\midrule
\multirow{3}{*}{DiffAffinity}
& all & \underline{0.661} & 0.544 & 1.536 & 1.101 & 0.742 & 0.422 & 0.392 \\
& single & \underline{0.668} & \underline{0.524} & \underline{1.279} & \underline{0.926} & \underline{0.731} & 0.440 & 0.413 \\
& multiple & 0.647 & 0.565 & 2.006 & 1.518 & \underline{0.776} & 0.376 & 0.351 \\
\midrule
\multirow{3}{*}{ProMIM}
& all & \textbf{0.672} & \textbf{0.573} & \underline{1.516} & \underline{1.089} & \textbf{0.760} & \textbf{0.464} & \textbf{0.431} \\
& single & \textbf{0.668} & \textbf{0.534} & \textbf{1.279} & \textbf{0.924} & \textbf{0.738} & \textbf{0.466} & \textbf{0.439} \\
& multiple & \textbf{0.666} & \textbf{0.614} & \textbf{1.963} & \textbf{1.491} & \textbf{0.825} & \textbf{0.458} & \underline{0.425} \\
\bottomrule
\end{tabular}
}
\end{table}
\endgroup

\subsection{Ablation Study and Analysis}
\label{sec:ablation}

\paragraph{Experimental Configuration}
To assess the effectiveness of different levels of interaction modeling, this section conducts an ablation study on the SKEMPI2 dataset. We follow the same 
experimental protocols in Section \ref{sec:skempi2}. Among all comparisons, ProMIM* refers to the same architecture but un-pretrained version of ProMIM.

\paragraph{Results}

As illustrated in Table \ref{tab:ablation}, missing any level of interaction modeling leads to suboptimal results, which indicates that each level of interaction modeling is indispensable for predicting the mutational effects. ProMIM outperforms the non-pre-trained ProMIM* by 8.29\% on the Spearman coefficient in per-complex evaluation, indicating that self-supervised learning from a large amount of data is very helpful for predicting mutational effects. Viewing each level of interaction modeling independently, we find that SIM performs the best, which confirms the argument that mutations mainly affect sidechain conformations. However, in predicting multi-point mutations, we found the best results came from PIM. This suggests that multi-point mutations may have a greater impact on side-chain conformation proportions compared to backbone conformation proportions, but ultimately result in significant changes in overall binding affinity.

\begingroup
\renewcommand{\arraystretch}{0.96} 

\begin{table}[t]
\small
\centering
\caption{Ablation study on the SKEMPI2 dataset. Results ranking first and second are highlighted in \textbf{bold} and \underline{underlined} respectively.}
\label{tab:ablation}
\resizebox{\textwidth}{!}{
\begin{tabular}{cl|rrrrr|rr}
\midrule
& & \multicolumn{5}{c|}{Overall} & \multicolumn{2}{c}{Per-Structure} \\
Method & Mutations & Pearson  & Spearman & RMSE & MAE & AUROC & Pearson & Spearman \\ 
\midrule
\multirow{3}{*}{ProMIM*}
& all & 0.637 & 0.534 & 1.578 & 1.127 & 0.732 & 0.443 & 0.398 \\
& single & 0.630 & 0.507 & 1.337 & 0.965 & 0.726 & 0.441 & 0.401 \\
& multiple & 0.646 & 0.578 & 2.013 & 1.499 & 0.762 & 0.433 & 0.379 \\
\midrule

\multirow{3}{*}{PIM}
& all & 0.649 & 0.544 & 1.556 & 1.115 & 0.743 & 0.422 & 0.381 \\
& single & 0.635 & 0.506 & 1.327 & 0.955 & 0.728 & 0.418 & 0.376 \\
& multiple & 0.666 & 0.606 & 1.963 & 1.487 & 0.788 & 0.430 & 0.402 \\
\midrule
\multirow{3}{*}{BIM}
& all & 0.623 & 0.528 & 1.597 & 1.152 & 0.739 & 0.426 & 0.395 \\
& single & 0.625 & 0.510 & 1.335 & 0.966 & \underline{0.732} & 0.427 & 0.397 \\
& multiple & 0.629 & 0.550 & 2.049 & 1.573 & 0.769 & 0.385 & 0.355 \\
\midrule
\multirow{3}{*}{SIM}
& all & 0.654 & 0.555 & 1.546 & \underline{1.104} & \underline{0.749} & 0.455 & 0.424 \\
& single & 0.647 & 0.515 & 1.307 & \underline{0.945} & 0.730 & 0.456 & 0.428 \\
& multiple & 0.647 & 0.596 & 2.005 & 1.506 & 0.806 & \underline{0.458} & \textbf{0.450} \\
\midrule
\multirow{3}{*}{PIM + BIM}
& all & 0.650 & 0.548 & 1.553 & 1.112 & 0.748 & 0.454 & 0.418 \\
& single & 0.641 & 0.508 & 1.313 & 0.951 & 0.731 & 0.439 & 0.395 \\
& multiple & \underline{0.667} & 0.611 & 1.965 & \textbf{1.480} & \underline{0.807} & 0.429 & 0.400 \\
\midrule

\multirow{3}{*}{PIM + SIM}
& all & 0.655 & 0.554 & 1.545 & 1.106 & 0.744 & \textbf{0.482} & \textbf{0.444} \\
& single & \underline{0.653} & \underline{0.520} & 1.298 & 0.941 & 0.731 & \textbf{0.476} & \textbf{0.441} \\
& multiple & 0.647 & 0.593 & 2.009 & 1.518 & 0.786 & 0.420 & 0.382 \\
\midrule

\multirow{3}{*}{BIM + SIM}
& all & \underline{0.662} & \underline{0.557} & \underline{1.533} & 1.108 & 0.743 & \underline{0.469} & 0.428 \\
& single & 0.650 & 0.518 & 1.304 & 0.949 & 0.726 & 0.456 & 0.419 \\
& multiple & \textbf{0.672} & \textbf{0.615} & \textbf{1.951} & \underline{1.485} & 0.799 & 0.454 & 0.409 \\
\midrule

\multirow{3}{*}{PIM + BIM + SIM}
& all & \textbf{0.672} & \textbf{0.573} & \textbf{1.516} &\textbf{1.089} & \textbf{0.760} & 0.464 & \underline{0.431} \\
& single & \textbf{0.668} & \textbf{0.534} & \textbf{1.279} & \textbf{0.924} & \textbf{0.738} & \underline{0.466} & \underline{0.439} \\
& multiple & 0.666 & \underline{0.614} & \underline{1.963} & 1.491 & \textbf{0.825} & \textbf{0.458} & \underline{0.425} \\
\midrule

\end{tabular}
}
\end{table}
\endgroup

\subsection{Sidechain Modeling is Not Enough for Mutational Effect Prediction}
\label{sec:sidechain_verify}

\paragraph{Experimental Configuration}
Previous advanced methods focusing on sidechain modeling, including RDE-Network \cite{luo2023rotamer} and DiffAffinity \cite{liu2023predicting}, identify their main limitation as the inability to model changes in backbone conformations upon mutation.
In addition to the previously discussed potential impact of multi-point mutations on conformational changes in Section~\ref{sec:skempi2} and Section~\ref{sec:ablation}, we aim to verify more directly whether this limitation exists for RDE-Network and DiffAffinity, and whether our ProMIM improves this issue through more comprehensive modeling of interactions. Previously, we have mentioned that mutations regarding glycine and proline can lead to significant changes in the backbone conformations \cite{rodrigues2019mcsm}. Therefore, we select test samples in Section~\ref{sec:ablation} that contain single-point mutations, from or to a glycine or proline, as our verification set.

\begin{wrapfigure}{r}{0.35\textwidth}



\centering
\includegraphics[width=\linewidth]{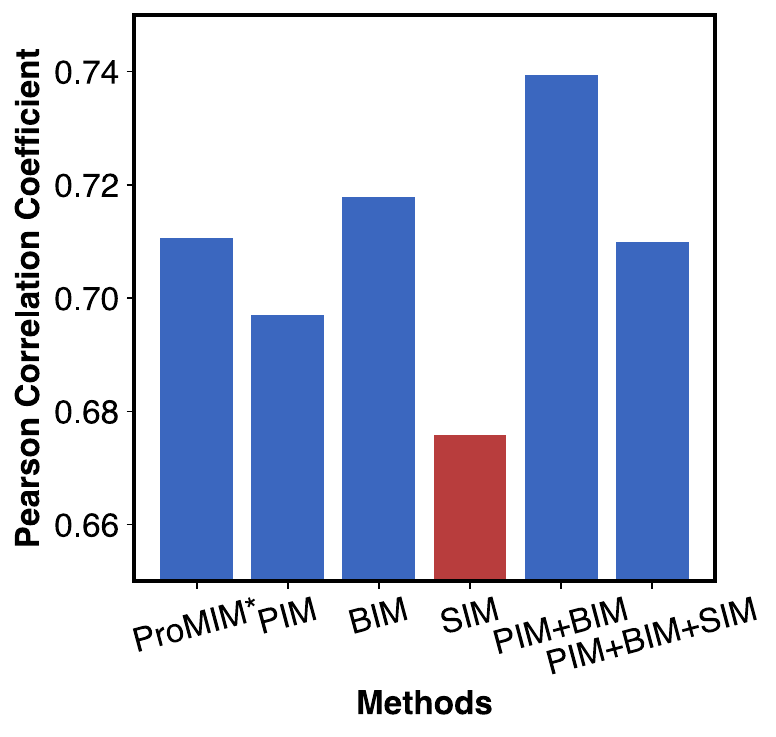}
\caption{An evaluation of $\Delta\Delta G$ prediction for backbone conformation-sensitive samples.
}
\label{fig:ablation_backbone}
    
\vspace{-10pt}

\end{wrapfigure}

\paragraph{Results}
From Figure \ref{fig:ablation_backbone}, SIM, which has the same setting as RDE-Network, performs the worst among all the methods, even significantly worse than the case without pre-training. This fully validates our hypothesis that merely modeling sidechains cannot accurately predict $\Delta \Delta G$ for mutations that involve changes in backbone conformations. In contrast, BIM shows a significant improvement over SIM, and the combination of PIM and BIM obtains the best result. This result proves that our interaction modeling in the protein and backbone levels indeed contributes to capturing the backbone conformation affected by mutations. However, combining all three levels of interaction modeling obtain sub-optimal performance. This does not mean we should abandon SIM, as its role is significant in cases where there are negligible changes in backbone conformations. Therefore, integrating interaction modeling at all three levels is a balanced choice for general cases, since we do not know whether the predicting target will involve changes in backbone conformations.


\begin{table}[ht]
\centering
\begin{minipage}[c]{.33\textwidth}
\centering
\small
\setlength{\tabcolsep}{5pt}
\caption{Evaluation of $\Delta\Delta G$ prediction for SARS-CoV-2 RBD.}


\label{tab:SARS ddg}
\begin{tabular}{lr}
\toprule
Method       & Pearson \\ 
\midrule
FoldX        & 0.385   \\
RDE-Net      & 0.438   \\
DiffAffinity & 0.466   \\ 
\midrule
ProMIM         & \textbf{0.483}   \\ 
\bottomrule
\end{tabular}
\end{minipage}
\hspace{6pt}
\begin{minipage}[c]{.63\textwidth}
\centering
\small
\setlength{\tabcolsep}{5pt}
\caption{
The ratio of the rankings of the five favorable mutation to the total number of mutations on the human antibody against SARS-CoV-2, lower is better. \textbf{Bold} values refer to ratio $\leq 10\%$.}
\label{tab:antibody}
\begin{tabular}{lrrrrr}
\toprule
Method       & TH31W  & AH53F   & NH57L   & RH103M  & LH104F  \\ 
\midrule
FoldX        & \textbf{4.25}\% & 14.57\% & \textbf{2.43}\%  & 27.13\% & 63.77\% \\
RDE-Net      & \textbf{5.06}\% & 12.15\% & 55.47\% & 50.61\% & \textbf{9.51}\%  \\
DiffAffinity & \textbf{7.28}\% & \textbf{3.64}\%  & 18.82\% & 81.78\% & 10.93\% \\ 
\midrule
ProMIM         & \textbf{5.33}\% & \textbf{4.79}\%  & 19.43\% & 75.78\% & \textbf{8.37}\%  \\ 
\bottomrule
\end{tabular}
\end{minipage}
\end{table}

\subsection{Prediction of Mutational Effects on Binding Affinity of SARS-CoV-2 RBD}
\label{sec:SARS ddg}
\paragraph{Experimental Configuration}
To clearly demonstrate ProMIM's generalization ability and potential for practical applications, we conducted two zero-shot experiments related to SARS-CoV-2. SARS-CoV-2 has received widespread attention and is not included in the existing mutation datasets. Our first experiment evaluates the zero-shot ability of mutational effect prediction. The testing data comes from a previous study~\cite{starr2022shifting}, which identified 15 significant mutation sites on the SARS-CoV-2 RBD that greatly influence its binding affinity with the ACE2 protein, and experimentally quantified the effects of 285 possible single-point mutations at these 15 sites using deep mutational scanning. We predict $\Delta\Delta G$ values for these 285 single-point mutations and calculate the Pearson correlation coefficient between the experimental and predicted $\Delta\Delta G$ values.

\paragraph{Results}
Since our experimental setting is the same as DiffAffintiy's \cite{liu2023predicting}, we take their baseline results. As Table \ref{tab:SARS ddg} shown, 
ProMIM shows a significant performance advantage over baselines, and since this is a zero-shot test, it indicates that ProMIM has stronger generalization ability for predicting mutational effects. This makes ProMIM more promising for use in practical applications to develop new therapies for diseases.

\subsection{Optimization of Human Antibodies against SARS-CoV-2}
\label{sec:antibody}
\paragraph{Experimental Configuration}
The second zero-shot experiment evaluates how well models identify favorable mutations on human antibodies. A previous study reported five favorable single-point mutations on a human antibody that can enhance neutralization efficacy against SARS-CoV-2~\cite{shan2022deep}. These mutations are among the 494 possible single-point mutations at 26 sites within the complementarity-determining region (CDR) of the antibody heavy chain. We use different methods to predict $\Delta\Delta G$ values for all 494 single-point mutations and rank them in ascending order, which means mutations with the lower predicted $\Delta\Delta G$ values will be ranked at the top to be more favorable mutations. Following the evaluation protocol of previous works \cite{luo2023rotamer,liu2023predicting}, we calculate the ratio of the rankings of the five favorable mutations to the total number of mutations. Lower values indicate better performance.

\paragraph{Results}
We employ the results of baselines presented in \cite{liu2023predicting} again for comparison. As indicated in Table \ref{tab:antibody}, ProMIM is the only method that successfully ranks 3 of the 5 favorable mutations within top 10\%, and identifies 4 of them within the top 20\% of the ranking. Developing new antibodies is of great significance for curing diseases, especially those with substantial social impact like SARS-CoV-2. ProMIM offers a promising approach for discovering new drugs.

\section{Conclusions}
\label{sec:conclusions}
In this work, we propose multi-level interaction modeling for protein mutational effect prediction (ProMIM) to fully model three levels of interaction that can be affected by mutation, including protein-level PIM, backbone-level BIM, and sidechain-level SIM. ProMIM achieves first and second place across all metrics in SKEMPI 2, demonstrating the effectiveness of modeling interactions at all three levels. Ablation studies show all three levels of interaction modeling are indispensable for mutational effect prediction. The overall pre-training objectives achieve 8.29\% relative improvement compared to the un-pretrained model, highlighting the importance of self-supervised learning for mutational effect prediction when labeled data is scarce. ProMIM's advancement over previous methods lies in its ability to model the impact of mutations on backbone conformational changes. This is evident from ProMIM's significant lead in multi-point mutations and the dataset sensitive to backbone structural changes. ProMIM's excellent performance in per-complex results and zero-shot experiments demonstrates its strong generalization ability and practical value. It has the potential to become a next-generation tool for developing new therapies and drugs. However, ProMIM still has considerable room for improvement in practical applications. For instance, if we are able to identify which mutations cause backbone conformational changes, we could potentially achieve better results with models focused more on backbone-level interaction modeling.



\bibliographystyle{unsrt}

\newpage
\appendix

\section{Implementation Details}
\label{apd:imp details}
\subsection{Pre-training Datasets}
During the pre-training stage, two datasets are used for different objectives. We employ the PPIRef50K dataset for the training of PIM and BIM, while the PDB-REDO dataset is utilized for SIM. This is because the training of PIM and BIM requires explicit and meaningful indication of the interacting binders, which is ambiguous in the PDB-REDO dataset. Although it is feasible to conduct SIM on the PPIRef50K dataset, using PPIRef50K for SIM could lead to reduced performance, as the amount of structures in the PPIRef50K is much smaller than that in the PDB-REDO. Besides, utilizing PDB-REDO for SIM makes it more explicit to demonstrate the effects of PIM and BIM in the comparison with baselines including RDE-Network and DiffAffinity. Detailed introductions of PPIRef50K \cite{bushuiev2023learning} and PDB-REDO \cite{joosten2014pdb_redo} are provided below.

\paragraph{PPIRef50K}
PPIRef50K is a non-redundant dataset of structurally distinct 3D protein-protein interfaces. These interfaces are selected from the Protein Data Bank as biophysically meaningful interactions under well-established criteria \cite{townshend2019end} and deduplicated to reduce structural redundancy. The resulting dataset comprises 45,553 PPIs, each of which involves the interaction between two protein chains.

\paragraph{PDB-REDO}
PDB-REDO is an extensive dataset containing over 130,000 refined X-ray structures in the Protein Data Bank. The protein chains in PDB-REDO are clustered based on 50\% sequence identity, resulting in 38,413 chain clusters.

\subsection{Training Details for Different Objectives}
Two encoders sharing the same architecture are trained on the PPIRef50K and PDB-REDO datasets independently with different pre-training objectives. The embedding sizes of single representations and pair representations are 128 and 64 respectively. 6 IPA blocks are used.

\paragraph{PIM and BIM} 
For a given protein-protein interface from the PPIRef50K dataset, we randomly select 64 residues from each of binders and form a structure of 128 residues as input. The losses of PIM and BIM are treated equally and trained in a multi-task manner. The model is optimized using the Adam optimizer for 200K iterations. The learning rate is set to 1e-4 initially and decays by 0.8 if the validation loss does not decrease in the last 5 validation steps. The minimum learning rate is 1e-6. We set the batch size to 48.

\paragraph{SIM}
We follow RDE \cite{luo2023rotamer} to preprocess the PDB-REDO dataset and use the same hyperparameters to train the model with the SIM objective. For each iteration, we randomly select a chain cluster from the PDB-REDO dataset and then randomly pick a chain from the cluster, following the same operation in RDE. Input structures are cropped into patches consisting of 128 residues.

For $\Delta\Delta G$ prediction, the hidden representations attained from the two pre-trained encoders are concatenated with the single representations and fused together using a one-layer MLP, which leads to updated single representations of size remaining 128. This operation is consistent with previous works including RDE-Network and DiffAffinity.

\section{Visualization of ProMIM's Performance on the SKEMPI2 Dataset}

\begin{figure}[htbp]
    \centering
    \begin{subfigure}[b]{0.325\textwidth}
        \centering
        \includegraphics[width=\textwidth]{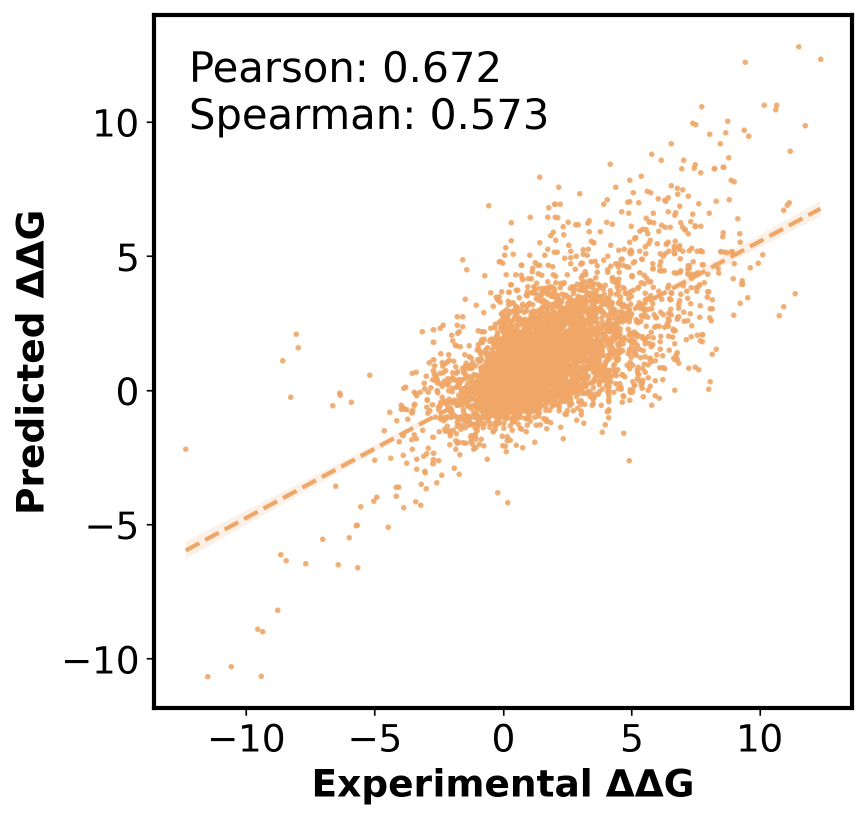}
        \caption{}
        \label{fig:promim_skempi2_all}
    \end{subfigure}
    \begin{subfigure}[b]{0.325\textwidth}
        \centering
        \includegraphics[width=\textwidth]{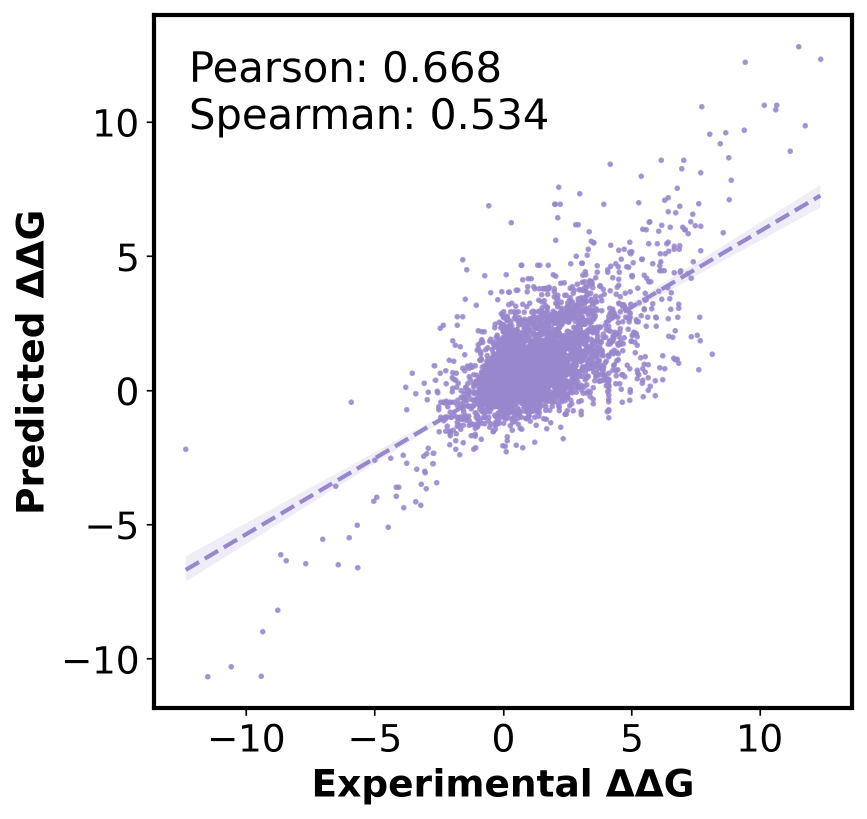}
        \caption{}
        \label{fig:promim_skempi2_single}
    \end{subfigure}
    \begin{subfigure}[b]{0.325\textwidth}
        \centering
        \includegraphics[width=\textwidth]{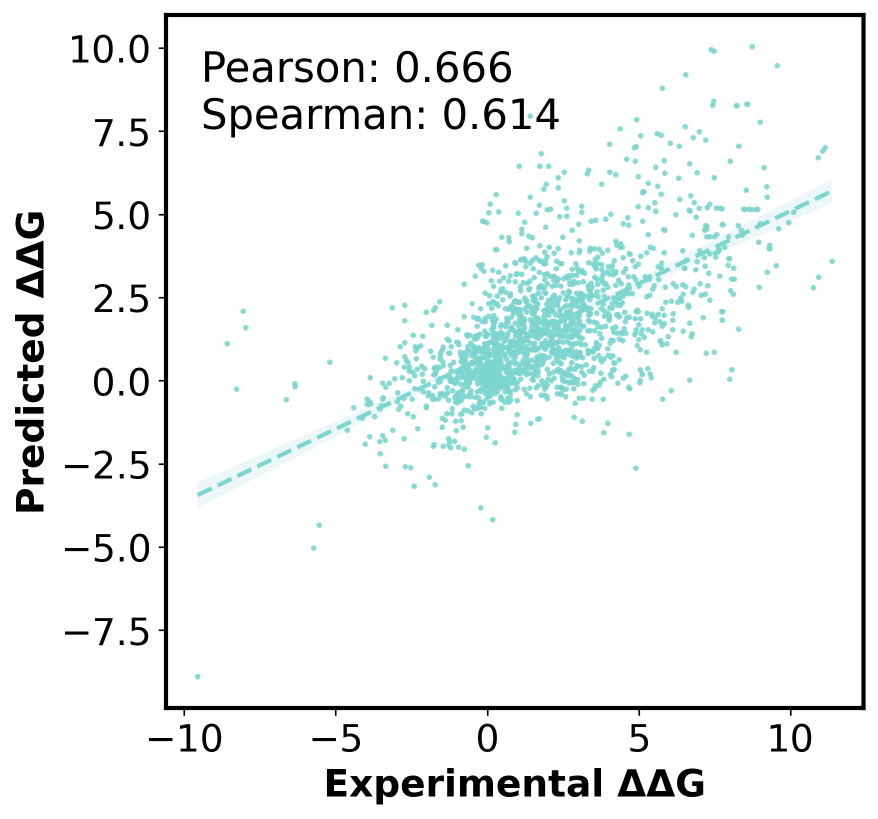}
        \caption{}
        \label{promim_skempi2_multiple}
    \end{subfigure}
    \caption{Performance of ProMIM on the SKEMPI2 dataset. (a) Correlation between experimental $\Delta\Delta G$s and $\Delta\Delta G$s predicted by ProMIM on the entire SKEMPI2 dataset. (b) Correlation between experimental $\Delta\Delta G$s and $\Delta\Delta G$s predicted by ProMIM on the SKEMPI2 single-mutation subset. (c) Correlation between experimental $\Delta\Delta G$s and $\Delta\Delta G$s predicted by ProMIM on the SKEMPI2 multi-mutation subset.}
    \label{fig:promim_skempi2}
\end{figure}

We visually demonstrate the performance of ProMIM on the SKEMPI2 dataset, as shown in Figure \ref{fig:promim_skempi2}.

\end{document}